\documentclass{article}
\usepackage[preprint]{neurips_2026}

\usepackage{amsmath,amssymb}
\usepackage{booktabs}
\usepackage{graphicx}
\usepackage{xcolor}
\usepackage[font=small,labelfont=bf]{caption}
\usepackage{hyperref}
\usepackage{microtype}
\usepackage{enumitem}
\usepackage{algorithm}
\usepackage{algpseudocode}
\usepackage{tcolorbox}
\tcbuselibrary{breakable,skins}
\usepackage{fvextra}

\usepackage{siunitx}  
\usepackage[table]{xcolor}

\definecolor{rowhl}{HTML}{EDF2F9}  
\definecolor{pgray}{HTML}{7B7B7B}   
\newcommand{\pv}[1]{\textcolor{pgray}{\scriptsize #1}}

\graphicspath{{paper_figures/}}

\definecolor{promptrule}{HTML}{2F6690}
\definecolor{promptback}{HTML}{F4F7FA}
\newtcolorbox{promptbox}[1]{%
  breakable, enhanced, sharp corners,
  colback=promptback, colframe=promptrule,
  boxrule=0.7pt, left=5pt, right=5pt, top=4pt, bottom=4pt,
  fonttitle=\bfseries\small, title={#1},
  before skip=8pt, after skip=8pt}

\newcommand{\promptsize}{\footnotesize}

\fvset{fontsize=\footnotesize, breaklines=true, breakanywhere=true,
       breaksymbolleft={}, xleftmargin=0pt}

\definecolor{revisionblue}{RGB}{0,70,180}
\definecolor{revisionpurple}{RGB}{125,0,160}

\title{Auditing Pairwise Equivalence Judgments: Self-Critique Effects and Diversity Measurement in Multi-Agent Hypothesis Generation}

\author{
  Ji Young Byun \\
  Johns Hopkins University \\
  Baltimore, MD 21218, USA \\
  {\footnotesize\texttt{jbyun13@jhu.edu}}
  \And
  Anthony Hu \\
  Merck \& Co., Inc. \\
  West Point, PA 19486, USA \\
  {\footnotesize\texttt{anthony.hu@merck.com}}
  \And
  Jesse Rogers \\
  Merck \& Co., Inc. \\
  West Point, PA 19486, USA \\
  {\footnotesize\texttt{jesse.rogers@merck.com}}
  \AND
  Roujia Wang \\
  Merck \& Co., Inc. \\
  West Point, PA 19486, USA \\
  {\footnotesize\texttt{roujia.wang@merck.com}}
  \And
  Manasa Kesapragada \\
  Merck \& Co., Inc. \\
  West Point, PA 19486, USA \\
  {\footnotesize\texttt{manasa.kesapragada@merck.com}}
  \And
  Falgun Shah \\
  Merck \& Co., Inc. \\
  West Point, PA 19486, USA \\
  {\footnotesize\texttt{falgun.shah@merck.com}}
}

\begin{document}
\maketitle
\footnotetext{Accepted to the Agentic AI for Biological Discovery Workshop (NeurIPS 2026).}

\begin{abstract}
Multi-agent systems built on large language models (LLMs) are increasingly applied to scientific discovery and hypothesis generation. Both the effect of refinement and the diversity of the delivered set are hard to interpret before experimental ground truth exists, and both are typically reported by deciding, for pairs of generated hypotheses, whether they describe the same underlying mechanism. We study two evaluation questions that rest on this pairwise equivalence judgment: (1) how much self-critique changes delivered hypotheses beyond run-to-run variability, and (2) how the equivalence rule used to group hypotheses affects measured diversity. Our study workflow generates mechanistic explanations represented as causal chains from a molecular initiating event through intermediate biological steps to an observed outcome, revised through self-critique rounds before a final consolidation step. Across four independent proprietary instances, we hold opening hypotheses fixed, rerun the downstream workflow with 0, 1, and 5 critique rounds, and score matched hypothesis pairs with an LLM-as-a-judge. Relative to matched same-depth reruns, moving from 0 to 1 round produces 34.5 percentage points (pp) of additional mechanism-level divergence, whereas 1 to 5 rounds adds 1.3 pp. We then compare three equivalence rules: term frequency--inverse document frequency (TF--IDF) similarity, dense embeddings, and the same LLM-as-a-judge. To characterize how these rules respond to specific edits, we construct controlled hypothesis pairs that either alter only wording or equivalent biological terminology while preserving the causal explanation, or replace one component of the causal chain while holding the remaining structure fixed. All three rules are invariant to meaning-preserving edits, but when the initiating event is replaced, the LLM-as-a-judge identifies 83\% of valid pairs as different mechanisms, versus 0\% for TF--IDF and 8\% for embeddings; varying only the rubric that defines same mechanism moves this figure from 38\% to 96\%. Together, these results show that pairwise equivalence judgments are a measurement choice: how mechanism equivalence is defined affects both the estimated effect of self-critique and the measured diversity of generated hypotheses.
\end{abstract}

\section{Introduction}
\label{sec:intro}

Large language models (LLMs) are increasingly being incorporated into scientific workflows, where agentic systems propose and refine hypotheses
\citep{gottweis2025cocientist,ghareeb2026robin}, interpret experimental data \citep{huang2025biomni, chen2025scienceagentbench}, design candidate molecules \citep{mbran2024chemcrow, swanson2025virtuallab}, and plan follow-up experiments \citep{huang2025biomni}.
Evaluating such systems departs from the standard benchmark setting. Scientific hypotheses often concern mechanisms whose truth is not yet established when they are proposed, and the experiments required to adjudicate them can be expensive and time-consuming. Correctness therefore becomes available, if at all, only after the system must already be evaluated, necessitating evaluation criteria that can be applied directly to the generated hypotheses.

In the absence of such an external signal, recent systems evaluate generated hypotheses by comparing them with one another, both while refining candidates and when reporting the delivered set. The AI co-scientist ranks hypotheses through an
Elo tournament of pairwise scientific debates, and separately consolidates its output set by clustering hypotheses in an embedding-based proximity graph \citep{gottweis2025cocientist}. Robin ranks candidates by asking an LLM-as-a-judge
which of two hypotheses is better across all pairs and fitting a Bradley--Terry--Luce model to the outcomes \citep{ghareeb2026robin}. \citet{si2025ideas} use both: generated ideas are deduplicated by thresholding pairwise embedding similarity before an LLM-as-a-judge orders the survivors. These systems apply different comparison rules to different purposes, with
LLM-based preference judgments prioritizing candidates and embedding similarity
supporting deduplication. 

Pairwise equivalence judgments provide a common measurement basis for two
evaluation questions that are otherwise treated separately. First, assessing
the effect of self-critique requires determining \emph{how much the delivered hypotheses change as critique depth varies}, relative to run-to-run variability across repeated executions at the same critique depth. Second, measuring the diversity of a generated set requires determining \emph{when two hypotheses should be treated as distinct explanations} rather than alternative formulations
of the same explanation. Although these questions concern different properties of the workflow, both involve the same hypothesis comparison task of deciding whether a pair should be treated as the same or different.

We study this problem in a multi-agent hypothesis generation workflow. Given an unexpected experimental finding, the workflow proposes mechanistic hypotheses that explain it, each represented as a causal chain from an initiating event through intermediate biological steps to that finding. For the first question, we hold the opening hypotheses fixed and vary only critique depth, then read cross-depth divergence against matched same-depth reruns, since repeating an identical configuration also changes the delivered hypotheses \citep{atil2024nondeterminism}. For the second, no labels exist for whether two generated hypotheses share a mechanism, so we compare equivalence rules on constructed pairs whose intended relation is known: some edits preserve the mechanism and should not change a rule's verdict, while others replace a component of the causal chain and should.

\begin{itemize}[leftmargin=1.5em,itemsep=0.25em,topsep=0.3em]

    \item We identify the pairwise equivalence judgment---deciding whether two
    hypotheses are the same or different---as a shared measurement basis for two
    evaluation questions that are usually treated separately: what changes beyond
    same-condition run-to-run variability, and what counts as one mechanism when
    measuring diversity.

    \item Replaying the downstream workflow from identical opening hypotheses, we
    show that the largest change in delivered mechanisms occurs when critique is
    first introduced: moving from 0 to 1 round produces 34.5 pp of divergence beyond
    matched rerun variability, whereas moving from 1 to 5 rounds adds 1.3 pp.

    \item We audit three equivalence rules with controlled perturbations. TF--IDF, embeddings, and an LLM-as-a-judge behave alike on meaning-preserving edits, but diverge when the initiating event is replaced: the LLM-as-a-judge reads 83\% of these pairs as different mechanisms, compared with 8\% for embeddings and 0\% for TF--IDF, and varying only the judge's rubric moves this rate from 38\% to
    96\%.

\end{itemize}

\section{Related Work}
\label{sec:related}

\paragraph{LLM agents for scientific hypothesis generation.}
LLM agents are now applied across the stages of a scientific workflow, from interpreting experimental data and designing candidate molecules to planning and executing experiments, and benchmarks have been built to measure these capabilities
\citep{boiko2023coscientist, mbran2024chemcrow, laurent2024labbench, huang2025biomni,swanson2025virtuallab, chen2025scienceagentbench}. Single-agent designs pursue it through iteration, alternating between generation and self-evaluation modules and reporting that self-evaluation improves the novelty and relevance of proposals
\citep{yang2025bioverge}. Multi-agent designs instead distribute the same work across specialized roles: the AI co-scientist separates generation, reflection, ranking, evolution, and proximity-based consolidation \citep{gottweis2025cocientist}, Robin couples literature-grounded generation with experimental analysis \citep{ghareeb2026robin}, and VirSci organizes a simulated research team that generates, discusses, and assesses ideas before selecting one \citep{su2025manyheads}.

\paragraph{Self-critique and iterative refinement.}
Refinement is a standard way to spend additional computation on a candidate output
before delivering it. A single agent can revise its own output from model-generated
feedback or environmental signals \citep{madaan2023selfrefine,shinn2023reflexion},
and multi-agent designs assign the critique to a separate instance that challenges
the proposal before a final response is produced \citep{du2024debate}. Test-time
scaling studies place this sequential revision alongside parallel sampling,
aggregation, and verifier-based merging as alternative ways to allocate the same
inference budget, both for LLMs and for the agents \citep{byun2026testtime,zhu2025agentscaling}. How much the added computation helps
varies with the setting: intrinsic self-correction without external feedback can
leave reasoning unimproved \citep{huang2024selfcorrect, kamoi2024selfcorrection} and
debate does not always outperform simpler ensembling \citep{smit2024mad}. In each
case the effect is read against a gold answer or a task-specific metric, a reference that open-ended hypothesis generation makes available only after the fact.

\paragraph{Comparative evaluation without a reference.}
When an external reference is available, a generation can be scored against it. FActScore decomposes text into atomic claims and assesses their evidential support, while ALCE evaluates the correctness and completeness of generated citations \citep{min2023factscore,gao2023alce}. Without such a reference, evaluation compares
generations with one another, and the common instrument is a pairwise LLM-as-a-judge that decides which of two outputs is better
\citep{zheng2023llmjudge,liusie2024comparative}. Because no reference is available for evaluating the  LLM-as-a-judge itself, responsible use requires understanding its behavior \citep{shi2024positionbias,eiras2025knowthyjudge}.

\section{Problem Formulation: Evaluation without Experimental Ground Truth}
\label{sec:prelim}

Given an unexpected experimental finding, a generation run delivers a set of
mechanistic hypotheses $\mathcal{H}=\{h_1,\ldots,h_n\}$ that could explain it, where
each $h_i$ is a causal chain from an initiating event through intermediate
biological steps to that finding, and $n$ varies across runs. Candidate hypotheses may be iteratively revised through self-critique before the final set is produced. Because the underlying biological mechanism is unknown at generation time, no experimental ground truth is available for deciding whether a proposed mechanism is correct. Such comparison requires an \emph{equivalence rule} $\rho_m(h_i,h_j)\in\{0,1\}$, indexed by the comparison method $m$, that returns $1$ when two hypotheses express the same mechanism.

\paragraph{What changes beyond run-to-run variability?}
Write $\mathcal{R}_K$ for the set of runs performed with up to $K$ critique rounds. For a run $r\in\mathcal{R}_K$, let $\mathcal{H}_r$ denote the set of hypotheses produced after self-critique. To compare two runs $r$ and $r'$, we form a set of matched hypothesis pairs $M(r,r')$, where each pair $(h,h')$ contains corresponding hypotheses from $\mathcal{H}_r$ and $\mathcal{H}_{r'}$. A blinded LLM-as-a-judge supplies the equivalence rule here, with $\rho_{\mathrm{LLM}}(h,h')\in\{0,1\}$ assigning $1$ when the pair describes the same mechanism and $0$ otherwise. We define the \emph{divergence} between the two runs as the fraction of matched hypothesis pairs judged to describe different mechanisms,

\begin{equation} 
D(r,r')= \frac{1}{|M(r,r')|} \sum_{(h,h')\in M(r,r')} \left[1-\rho_{\mathrm{LLM}}(h,h')\right]. 
\label{eq:divergence}
\end{equation}

Here, $|M(r,r')|$ is the number of matched hypothesis pairs used to compare the two runs. Thus, $D(r,r')=0$ when all matched pairs are judged to describe the same mechanism, while larger values indicate greater mechanism-level divergence between the two runs.

A cross-depth difference alone does not indicate how much of the observed change is specific to critique depth, because rerunning the same configuration can also produce different hypotheses. Let $\bar{D}_{K,K'}$ denote the mean divergence over run pairs with critique depths $K$ and $K'$, and let $\bar{D}^{\mathrm{within}}_{K,K'}$ denote the mean divergence over all same-depth run pairs from the two configurations. We define the \emph{extra divergence} as

\begin{equation}
\Delta(K,K') = \bar{D}_{K,K'} - \bar{D}^{\mathrm{within}}_{K,K'}, \label{eq:extra_divergence}
\end{equation}

with the opening hypotheses held fixed so that critique depth is the only experimentally varied configuration. A positive $\Delta(K,K')$ indicates that runs at different critique depths diverge more than expected from rerunning the corresponding same-depth configurations, whereas $\Delta(K,K')=0$ indicates no additional divergence beyond this run-to-run variability.

\paragraph{What counts as one mechanism when measuring diversity?}
Once hypotheses have been generated, reporting their diversity requires grouping
$\mathcal{H}$ into distinct mechanisms and counting the resulting groups. Here $m$ ranges
over the candidate comparison methods---TF--IDF similarity, dense embeddings, and an LLM-as-a-judge---and the same grouping procedure is applied to the pairwise decisions of each,
\begin{equation}
    \widehat{P}_m=\mathcal{P}\!\left(\mathcal{H};\rho_m\right),
    \label{eq:method_dependence}
\end{equation}
where $\mathcal{P}$ clusters $\mathcal{H}$ under $\rho_m$ and returns the number of
resulting clusters. Because no reference is available for scoring $\rho_m$ either, we
audit the candidate rules with targeted perturbations: some preserve the underlying
mechanism, whereas others replace a specified component of the causal chain.

\section{Experimental Setup}
\label{sec:settings}

\begin{figure}[t]
\centering
\includegraphics[width=0.80\linewidth]{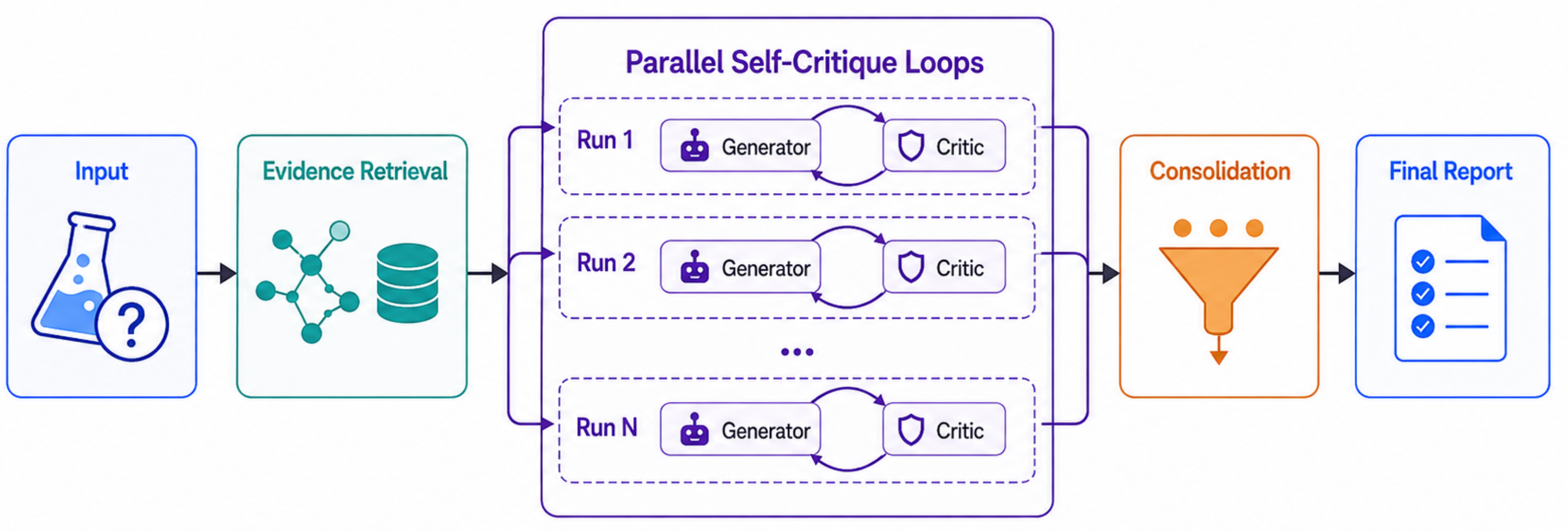}
\caption{\textbf{Multi-agent workflow for mechanistic hypothesis generation used in
this study.} An unexpected experimental finding initiates evidence retrieval. Candidate hypotheses are generated in parallel and iteratively refined through self-critique loops between proposer and critic agents. The resulting hypotheses are then consolidated into a final report.}
\label{fig:workflow}
\end{figure}

We study our multi-agent workflow on four independent proprietary instances, each
defined by an unexpected experimental finding and its retrieved evidence (Figure~\ref{fig:workflow}). The instances are proprietary and are anonymized as A--D throughout. For each instance, we perform five independent hypothesis-generation and refinement processes, using retrieved evidence and up to
$K$ rounds of self-critique before consolidation.

Generation and evaluation are kept separate: all evaluation is performed post hoc and does not influence hypothesis generation or refinement. Generation uses \texttt{Claude Sonnet 4.6} for the proposer and critic. Two judges are used post hoc, both \texttt{Claude Opus 4.6} at temperature~0: one labels how the proposer responds to each critique, and one applies the equivalence rule that decides whether two hypotheses describe the same mechanism. Dense representations use \texttt{Amazon Titan Text Embeddings V2}. The equivalence judgments are additionally repeated across three model families (\texttt{Claude Opus 4.6}, \texttt{Claude Sonnet 4.6}, \texttt{GPT-5.5}) to assess sensitivity. These models are used exclusively for post-hoc evaluation and provide no feedback to the generation process. Appendix~\ref{app:methods} provides the details.

\section{Results}
\label{sec:results}

We first examine how changing critique depth alters the delivered mechanisms relative to the run-to-run variability observed under the same critique setting. We then examine how different equivalence rules affect measured diversity by changing which generated hypotheses are treated as distinct mechanisms.

\subsection{Critique-depth effects relative to run-to-run variability}
\label{sec:adversarial}

\paragraph{Critique is actively incorporated during refinement.}
Across the critique exchanges from the four proprietary instances, a blinded post-hoc LLM-as-a-judge decomposes the critiques into 332 individual demands and finds that 61\% are conceded, 27\% are answered, 9\% are partially addressed, and 3\% are dropped. These results show that the refinement process engages with the critiques, but do not establish how much critique changes the final hypotheses or whether those changes are scientifically beneficial.

\paragraph{Most critique-depth divergence appears after the first critique round.}
Holding the opening hypotheses fixed, we replay the downstream process with 0, 1, and 5 critique rounds and compare the resulting hypotheses against matched same-depth reruns. All comparisons are conducted within instance to account for differences in the number and composition of generated hypotheses. The blinded LLM-as-a-judge is not given the critique depth or provenance of either hypothesis.

\begin{figure}[t]
\centering
\includegraphics[width=0.85\linewidth]{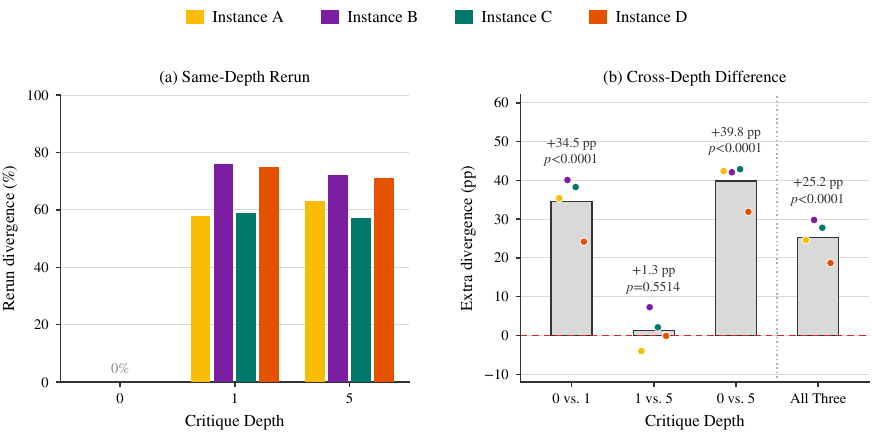}
\caption{\textbf{Effect of critique depth relative to run-to-run variability.} (a) Divergence between same-depth reruns at 0, 1, and 5 critique rounds.
(b) Extra divergence when critique depth changes, measured relative to the corresponding same-depth rerun variability. Most of the cross-depth separation occurs when critique is first introduced (0 to 1 round), with little additional separation from 1 to 5 rounds. Points show individual instances and bars show the aggregate result across instances.}
\label{fig:rq1main}
\end{figure}

Under the primary evaluator shown in Figure~\ref{fig:rq1main}, same-depth reruns with no critique show 0\% divergence for all four instances, whereas reruns with 1 critique round show 58\%--76\% divergence across instances. Although downstream generation and consolidation remain stochastic, the zero-round condition is substantially reproducible at the mechanism level. Relative to the corresponding same-depth baselines, moving from 0 to 1 critique round produces 34.5 percentage points (pp) of extra divergence, whereas moving from 1 to 5 rounds adds 1.3 pp. Moving from 0 to 5 rounds produces 39.8 pp of extra divergence. Thus, most of the cross-depth separation appears when critique is first introduced, with little additional separation from later rounds. This pattern is robust to evaluator choice: across three evaluator configurations, the 0-to-1 extra divergence ranges from 34.5 to 37.3 pp, while the corresponding same-depth rerun ranges are 0\%--5\% at zero critique and 58\%--76\% at one critique round (Appendix~\ref{app:capablation}).

The preceding divergence estimates were obtained under a single equivalence
rule, implemented by an LLM-as-a-judge  that determined whether each matched pair described the same mechanism. Pairwise equivalence judgments also underlie the measurement of output diversity. When hypotheses are generated in parallel, estimating diversity requires grouping hypotheses judged to express the same explanation, and the resulting number of distinct groups may vary with the equivalence rule used. We therefore compare three equivalence rules applied to
the same set of hypotheses.

\subsection{Measured diversity depends on the equivalence rule}
\label{sec:diversity}

We compare three rules for deciding whether two generated hypotheses describe the same mechanism. TF--IDF similarity scores a pair by weighted lexical overlap, so that shared rare terms count for more than shared common ones
\citep{sparckjones1972idf}. Dense embedding similarity replaces this surface match
with cosine distance between learned sentence representations, which can align
paraphrases that share little vocabulary \citep{reimers2019sbert}. An LLM-as-a-judge is shown both hypotheses and returns a same-or-different verdict under a written rubric \citep{zheng2023llmjudge}. We examine
the pairwise decisions directly, using controlled perturbations that either preserve the underlying mechanism or alter a specified component of the causal chain.

\paragraph{Targeted perturbations probe what each rule treats as a distinct mechanism.}
Each hypothesis contains a natural language mechanistic explanation and a causal chain from an initiating event through intermediate biological steps to the observed findings. We construct four controlled variants, each changing a specific aspect of the original hypothesis. \textit{Wording Only} rewrites the mechanistic explanation while leaving the causal chain unchanged. \textit{Wording + Chain} additionally rewrites the chain using audited biological aliases while preserving the same entities and causal structure. Both are designed to preserve the underlying mechanism. \textit{First Step Swapped} replaces the initiating event while keeping the downstream structure fixed and is designed to represent a different mechanism under our study definition. \textit{Middle Step Swapped} replaces one intermediate causal step while preserving the initiating event and terminal finding. Details are included in Appendix~\ref{app:variants}.

\begin{figure}[t]
\centering

\begin{minipage}[t]{0.49\linewidth}
\vspace{0pt}
\centering
\includegraphics[width=\linewidth]{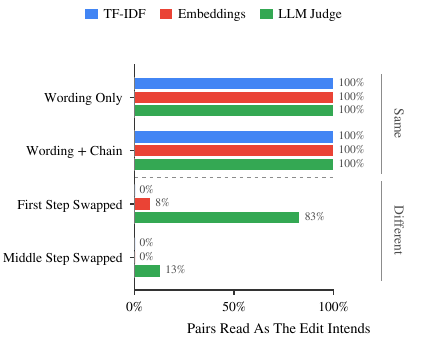}
\captionof{figure}{\textbf{Responses to targeted perturbations.}
All three rules classify \textit{Wording Only} and \textit{Wording + Chain} variants as the same mechanism. When the causal chain changes: the LLM-as-a-judge is more sensitive to \textit{First Step Swapped} changes than TF--IDF or embeddings, while all three are less sensitive to \textit{Middle Step Swapped} changes.}
\label{fig:arms}
\end{minipage}
\hfill
\begin{minipage}[t]{0.49\linewidth}
\vspace{0pt}
\centering
\includegraphics[width=\linewidth]{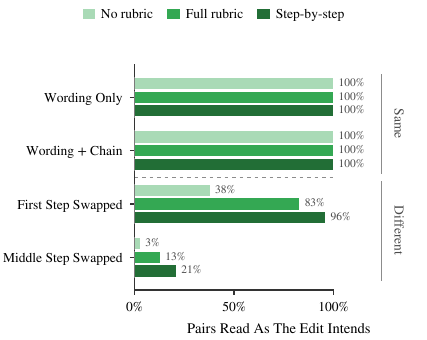}
\captionof{figure}{\textbf{Prompt specification changes the causal differences detected by the LLM-as-a-judge.} All three prompts remain invariant to \textit{Wording Only} and \textit{Wording + Chain} edits, while more explicit criteria increase sensitivity to causal changes: for \textit{First Step Swapped}, the share classified as different rises from 38\% with no rubric to 83\% with the full rubric and 96\% with the step-by-step rubric, which also increases sensitivity to \textit{Middle Step Swapped} changes.}
\label{fig:judgeprompts}
\end{minipage}

\end{figure}

\paragraph{Targeted perturbations reveal different rule sensitivities.}
All three rules classify every valid \textit{Wording Only} and \textit{Wording + Chain} pair as the same mechanism, indicating robustness to paraphrasing and audited alias substitutions (Figure~\ref{fig:arms}). Differences emerge when the causal structure is altered. For \textit{First Step Swapped}, the LLM-as-a-judge classifies 83\% of pairs as different, compared with 0\% for TF--IDF and 8\% for embeddings, suggesting that global similarity measures may be less sensitive when a single causally important component changes while most of the hypothesis remains unchanged. For \textit{Middle Step Swapped}, the LLM-as-a-judge classifies only 13\% of pairs as different, while TF--IDF and embeddings classify none as different, indicating that even the LLM-as-a-judge often treats changes to intermediate causal steps as variation within the same mechanism. This motivates examining how the LLM-as-a-judge's decisions change when the evaluation rubric is specified differently.

\paragraph{LLM-as-a-judge behavior depends on prompt specification.}
We test whether the LLM-as-a-judge's same or different decisions depend on how the evaluation criteria are specified. We compare three prompt variants: the \emph{full rubric} used throughout the paper, a \emph{step-by-step} rubric that separately evaluates the initiating event, causal route, and terminal outcome before the overall verdict, and a \emph{no-rubric} version that provides only the comparison task and verdict field. All prompts are provided in Appendix~\ref{app:prompts}.

\paragraph{Explicit criteria improve sensitivity to causal differences.}
In Figure~\ref{fig:judgeprompts}, all three prompts remain invariant to \textit{Wording Only} and \textit{Wording + Chain} edits. The largest difference appears for \textit{First Step Swapped}: the no-rubric prompt classifies 38\% of pairs as different, compared with 83\% under the full rubric and 96\% under the step-by-step rubric. The step-by-step rubric also increases sensitivity to \textit{Middle Step Swapped} changes, from 13\% under the full rubric to 21\%, while explicitly separating judgments of the initiating event, causal route, and terminal outcome. Reversing the presentation order changes decisions for 5.6\% of controlled perturbation pairs under each prompt variant, although the affected controlled pairs differ across rubrics (Appendix~\ref{app:stability}). Together, these results suggest that making the comparison criteria more explicit helps the LLM-as-a-judge distinguish causally meaningful differences while preserving invariance to meaning-preserving edits.

\section{Conclusion}
\label{sec:conclusion}
We studied two questions for evaluating scientific hypothesis generation before
experimental ground truth is available: how much self-critique changes delivered
mechanisms beyond run-to-run variability, and how the equivalence rule affects
measured diversity. Moving from 0 to 1 critique round produced 34.5 pp of extra
divergence, whereas 1 to 5 rounds added 1.3 pp. Under controlled perturbations, all
three rules were robust to meaning-preserving edits but differed on causal changes:
for \textit{First Step Swapped}, the LLM-as-a-judge read 83\% of pairs as different,
compared with 8\% for embeddings and 0\% for TF--IDF, and this rate varied from 38\%
to 96\% with the rubric alone. These results suggest that evaluation should account for variability and carefully consider both the equivalence method and the rubric used to interpret measured diversity.

\paragraph{Limitations.}
Our study is limited to four proprietary instances and five repetitions per critique depth, and no biological ground truth is available for the generated hypotheses. The results therefore characterize how the workflow outputs and evaluation methods behave, rather than whether critique improves scientific correctness or whether a particular grouping corresponds to the biologically correct mechanism.

Our post-hoc evaluation also relies on repeated LLM-based comparisons, which may become computationally expensive as the number of generated hypotheses increases. Pairwise mechanism comparisons grow rapidly with the size of the hypothesis set, while robustness analyses across judge models and prompt variants introduce additional inference cost.

Finally, measured diversity remains sensitive to evaluator design. TF--IDF and embedding results depend on similarity thresholds, while LLM same or different judgments vary with judge model, prompt specification, and presentation order (Appendix~\ref{app:stability}). Because the judges and generators belong to the same broad class of LLMs, agreement across judge families does not rule out shared biases such as self-preference or sensitivity to superficial properties. Our controlled perturbations also probe only a limited set of semantic and causal changes. Broader, systematically constructed perturbation suites and independently specified evaluation rubrics would provide stronger evidence about which causal distinctions an evaluator captures and how robustly it captures them.

\paragraph{Future work.}
A first direction is to improve the efficiency of post-hoc evaluation, for example through selective pairwise comparison, cascaded evaluators that invoke an LLM-as-a-judge only for ambiguous pairs, or distillation of LLM judgments into smaller models. A second direction is to strengthen the evaluation reference itself. Blinded domain experts could help validate controlled perturbations, refine the rubric for distinguishing mechanisms, and adjudicate same/different decisions. Measuring expert--expert and expert--model agreement across a broader range of mechanistic differences would help identify where automated evaluators are reliable and where their judgments remain ambiguous. Ultimately, prospective experimental evidence will be needed to determine whether critique-induced changes and evaluator-defined distinctions correspond to scientifically meaningful biological mechanisms.

\clearpage
\begingroup
\small
\bibliographystyle{unsrtnat}
\bibliography{references}

@misc{gottweis2025cocientist,
  author        = {Gottweis, J. and Weng, W.-H. and Daryin, A. and others},
  title         = {Towards an {AI} Co-Scientist},
  year          = {2025},
  archiveprefix = {arXiv},
  eprint        = {2502.18864}
}

@article{ghareeb2026robin,
  author  = {Ghareeb, A. E. and Chang, B. and Mitchener, L. and others},
  title   = {A Multi-Agent System for Automating Scientific Discovery},
  journal = {Nature},
  volume  = {655},
  pages   = {497--505},
  year    = {2026}
}

@inproceedings{yang2025bioverge,
  author    = {Yang, F. and Ye, C. and Ma, M. D. and Xiao, Y. and Yang, M. and Wang, W.},
  title     = {{BioVerge}: A Comprehensive Benchmark and Study of Self-Evaluating Agents for Biomedical Hypothesis Generation},
  booktitle = {NeurIPS Workshop on AI for Science},
  year      = {2025}
}

@inproceedings{madaan2023selfrefine,
  author    = {Madaan, A. and Tandon, N. and Gupta, P. and others},
  title     = {{Self-Refine}: Iterative Refinement with Self-Feedback},
  booktitle = {Advances in Neural Information Processing Systems},
  year      = {2023}
}

@inproceedings{shinn2023reflexion,
  author    = {Shinn, N. and Cassano, F. and Gopinath, A. and Narasimhan, K. and Yao, S.},
  title     = {Reflexion: Language Agents with Verbal Reinforcement Learning},
  booktitle = {Advances in Neural Information Processing Systems},
  year      = {2023}
}

@inproceedings{du2024debate,
  author    = {Du, Y. and Li, S. and Torralba, A. and Tenenbaum, J. B. and Mordatch, I.},
  title     = {Improving Factuality and Reasoning in Language Models through Multiagent Debate},
  booktitle = {Proceedings of the 41st International Conference on Machine Learning},
  year      = {2024}
}

@inproceedings{smit2024mad,
  author    = {Smit, A. P. and Grinsztajn, N. and Duckworth, P. and Barrett, T. D. and Pretorius, A.},
  title     = {Should We Be Going {MAD}? A Look at Multi-Agent Debate Strategies for {LLM}s},
  booktitle = {Proceedings of the 41st International Conference on Machine Learning},
  year      = {2024}
}

@inproceedings{min2023factscore,
  author    = {Min, S. and Krishna, K. and Lyu, X. and others},
  title     = {{FActScore}: Fine-Grained Atomic Evaluation of Factual Precision in Long-Form Text Generation},
  booktitle = {Proceedings of EMNLP},
  year      = {2023}
}

@inproceedings{gao2023alce,
  author    = {Gao, T. and Yen, H. and Yu, J. and Chen, D.},
  title     = {Enabling Large Language Models to Generate Text with Citations},
  booktitle = {Proceedings of EMNLP},
  year      = {2023}
}

@inproceedings{zheng2023llmjudge,
  author    = {Zheng, L. and Chiang, W.-L. and Sheng, Y. and others},
  title     = {Judging {LLM}-as-a-Judge with {MT-Bench} and {Chatbot Arena}},
  booktitle = {Advances in Neural Information Processing Systems},
  year      = {2023}
}

@inproceedings{su2025manyheads,
title     = {Many Heads Are Better Than One: Improved Scientific Idea
Generation by A {LLM}-Based Multi-Agent System},
author    = {Su, Haoyang and Chen, Renqi and Tang, Shixiang and Yin, Zhenfei
and Zheng, Xinzhe and Li, Jinzhe and Qi, Biqing and Wu, Qi
and Li, Hui and Ouyang, Wanli and Torr, Philip and Zhou, Bowen
and Dong, Nanqing},
booktitle = {Proceedings of the 63rd Annual Meeting of the Association
for Computational Linguistics},
pages     = {28201--28240},
year      = {2025},
doi       = {10.18653/v1/2025.acl-long.1368}
}

@article{shi2024positionbias,
title   = {Judging the Judges: A Systematic Study of Position Bias in
{LLM}-as-a-Judge},
author  = {Shi, Lin and Ma, Chiyu and Liang, Wenhua and Ma, Weicheng
and Vosoughi, Soroush},
journal = {arXiv preprint arXiv:2406.07791},
year    = {2024}
}

@inproceedings{byun2026testtime,
  title     = {Test-Time Scaling in Clinical Decision Making},
  author    = {Byun, Ji Young and Park, Young-Jin and
               Azizan, Navid and Chellappa, Rama},
  booktitle = {Proceedings of the 9th International Conference on
               Medical Imaging with Deep Learning},
  series    = {Proceedings of Machine Learning Research},
  volume    = {315},
  pages     = {2118--2142},
  year      = {2026},
  publisher = {PMLR},
  url       = {https://proceedings.mlr.press/v315/byun26a.html}
}

@article{swanson2025virtuallab,
  author  = {Kyle Swanson and Wesley Wu and Nash L. Bulaong and John E. Pak and James Zou},
  title   = {The {V}irtual {L}ab of {AI} agents designs new {SARS}-{C}o{V}-2 nanobodies},
  journal = {Nature},
  year    = {2025},
  doi     = {10.1038/s41586-025-09442-9}
}

@article{huang2025biomni,
  author  = {Kexin Huang and others},
  title   = {Biomni: A General-Purpose Biomedical {AI} Agent},
  journal = {bioRxiv},
  year    = {2025},
  doi     = {10.1101/2025.05.30.656746}
}

@article{boiko2023coscientist,
  author  = {Daniil A. Boiko and Robert MacKnight and Ben Kline and Gabe Gomes},
  title   = {Autonomous chemical research with large language models},
  journal = {Nature},
  volume  = {624},
  number  = {7992},
  pages   = {570--578},
  year    = {2023},
  doi     = {10.1038/s41586-023-06792-0}
}

@article{mbran2024chemcrow,
  author  = {Andres M. Bran and Sam Cox and Oliver Schilter and Carlo Baldassari and Andrew D. White and Philippe Schwaller},
  title   = {Augmenting large language models with chemistry tools},
  journal = {Nature Machine Intelligence},
  volume  = {6},
  pages   = {525--535},
  year    = {2024},
  doi     = {10.1038/s42256-024-00832-8}
}

@inproceedings{huang2024selfcorrect,
  author    = {Jie Huang and Xinyun Chen and Swaroop Mishra and Huaixiu Steven Zheng
               and Adams Wei Yu and Xinying Song and Denny Zhou},
  title     = {Large Language Models Cannot Self-Correct Reasoning Yet},
  booktitle = {International Conference on Learning Representations},
  year      = {2024}
}

@article{kamoi2024selfcorrection,
  author  = {Ryo Kamoi and Yusen Zhang and Nan Zhang and Jiawei Han and Rui Zhang},
  title   = {When Can {LLM}s Actually Correct Their Own Mistakes?
             A Critical Survey of Self-Correction of {LLM}s},
  journal = {Transactions of the Association for Computational Linguistics},
  volume  = {12},
  pages   = {1417--1440},
  year    = {2024}
}

@inproceedings{si2025ideas,
  author    = {Chenglei Si and Diyi Yang and Tatsunori Hashimoto},
  title     = {Can {LLM}s Generate Novel Research Ideas?
               A Large-Scale Human Study with 100+ {NLP} Researchers},
  booktitle = {International Conference on Learning Representations},
  year      = {2025}
}

@article{atil2024nondeterminism,
  author  = {Berk Atil and Alexa Chittams and Liseng Fu and Ferhan Ture and Lixinyu Xu and Breck Baldwin},
  title   = {Non-Determinism of ``Deterministic'' {LLM} Settings},
  journal = {arXiv preprint arXiv:2408.04667},
  year    = {2024}
}

@article{laurent2024labbench,
  author  = {Jon M. Laurent and Joseph D. Janizek and Michael Ruzo and
             Michaela M. Hinks and Michael J. Hammerling and Siddharth Narayanan and
             Manvitha Ponnapati and Andrew D. White and Samuel G. Rodriques},
  title   = {{LAB-Bench}: Measuring Capabilities of Language Models for Biology Research},
  journal = {arXiv preprint arXiv:2407.10362},
  year    = {2024}
}

@inproceedings{chen2025scienceagentbench,
  author    = {Ziru Chen and Shijie Chen and Yuting Ning and Qianheng Zhang and
               Boshi Wang and Botao Yu and Yifei Li and Zeyi Liao and Chen Wei and
               Zitong Lu and Vishal Dey and Mingyi Xue and Frazier N. Baker and
               Benjamin Burns and Daniel Adu-Ampratwum and Xuhui Huang and
               Xia Ning and Song Gao and Yu Su and Huan Sun},
  title     = {{ScienceAgentBench}: Toward Rigorous Assessment of Language Agents
               for Data-Driven Scientific Discovery},
  booktitle = {International Conference on Learning Representations},
  year      = {2025}
}

@article{zhu2025agentscaling,
  author  = {King Zhu and Hanhao Li and Siwei Wu and Tianshun Xing and Dehua Ma and
             Xiangru Tang and Minghao Liu and Jian Yang and Jiaheng Liu and
             Yuchen Eleanor Jiang and Changwang Zhang and Chenghua Lin and
             Jun Wang and Ge Zhang and Wangchunshu Zhou},
  title   = {Scaling Test-time Compute for {LLM} Agents},
  journal = {arXiv preprint arXiv:2506.12928},
  year    = {2025}
}

@article{sparckjones1972idf,
  author  = {Karen Sp{\"a}rck Jones},
  title   = {A Statistical Interpretation of Term Specificity and Its Application
             in Retrieval},
  journal = {Journal of Documentation},
  volume  = {28},
  number  = {1},
  pages   = {11--21},
  year    = {1972}
}

@inproceedings{reimers2019sbert,
  author    = {Nils Reimers and Iryna Gurevych},
  title     = {{Sentence-BERT}: Sentence Embeddings using Siamese {BERT}-Networks},
  booktitle = {Proceedings of the 2019 Conference on Empirical Methods in
               Natural Language Processing},
  pages     = {3982--3992},
  year      = {2019}
}

@inproceedings{liusie2024comparative,
  author    = {Adian Liusie and Potsawee Manakul and Mark Gales},
  title     = {{LLM} Comparative Assessment: Zero-Shot {NLG} Evaluation through
               Pairwise Comparisons using Large Language Models},
  booktitle = {Proceedings of the 18th Conference of the European Chapter of the
               Association for Computational Linguistics},
  pages     = {139--151},
  year      = {2024}
}

@article{eiras2025knowthyjudge,
  author  = {Francisco Eiras and Eliott Zemour and Eric Lin and Vaikkunth Mugunthan},
  title   = {Know Thy Judge: On the Robustness Meta-Evaluation of {LLM} Safety Judges},
  journal = {arXiv preprint arXiv:2503.04474},
  year    = {2025}
}
\endgroup

\clearpage
\appendix

\section{Critique-Depth Effects Relative to Run-to-Run Variability}
\label{app:capablation}

This analysis quantifies how much changing critique depth alters the delivered mechanisms beyond the run-to-run variability observed when the same critique setting is repeated. Holding the opening hypotheses fixed, we replay the downstream workflow with 0, 1, and 5 critique rounds and use a blinded LLM-as-a-judge to compare matched delivered hypotheses. For each run pair, divergence $D(r,r')$ is the proportion of matched hypothesis pairs classified as different mechanisms. Extra divergence $\Delta(K,K')$ is the mean cross-depth divergence minus the mean divergence between same-depth reruns at the corresponding depths. Formal definitions are provided in Section~\ref{sec:prelim}.

For each of instances A--D, one donor run provides five opening hypotheses, which are held fixed across all replays while every downstream stage is rerun. Each critique depth is repeated five times per instance, yielding 15 runs per instance and 60 runs in total. The 15 runs produce 105 run-pair comparisons per instance and 420 across all four instances, comprising 4{,}863 matched hypothesis pairs.

For each instance and each pairwise depth comparison, the five runs from each depth are pooled, and all 126 unique partitions of the ten runs into two groups of five are considered. We recompute $\Delta$ for each partition, and the two-sided $p$ value is the proportion of partitions that produce an absolute $\Delta$ at least as large as the observed value. For the aggregate result, this permutation is performed separately within each instance, after which the four instance-specific $\Delta$ values are averaged. 

\begin{table}[h]
\centering
\caption{\textbf{Cross-depth separation is concentrated in comparisons between the no-critique condition and conditions with critique.} Instance-averaged extra divergence, measured relative to the corresponding same-depth rerun divergence, is 34.5,pp for $K=0$ versus $K=1$ and 39.8,pp for $K=0$ versus $K=5$, compared with 1.3,pp for $K=1$ versus $K=5$. Gray entries report $p$ values.}

\label{tab:capablation}

\small
\setlength{\tabcolsep}{5.5pt}
\renewcommand{\arraystretch}{1.05}

\begin{tabular}{
    @{}l
    *{3}{S[table-format=2.0]}
    *{4}{S[table-format=+2.1, print-implicit-plus]}
    @{}
}
\toprule
& \multicolumn{3}{c}{\textbf{Same-Depth Rerun}}
& \multicolumn{4}{c}{\textbf{Cross-Depth Difference}} \\
\cmidrule(lr){2-4}
\cmidrule(l){5-8}

& \multicolumn{3}{c}{\footnotesize Divergence (\%)}
& \multicolumn{4}{c}{\footnotesize Extra divergence (pp)} \\

Instance
& {$K=0$}
& {$K=1$}
& {$K=5$}
& {$0\!\rightarrow\!1$}
& {$1\!\rightarrow\!5$}
& {$0\!\rightarrow\!5$}
& {All Three} \\
\midrule

Instance A
& 0 & 58 & 63
& +35.4 & -4.0 & +42.3 & +24.6 \\[-2pt]
& & & &
{\pv{.008}} &
{\pv{.556}} &
{\pv{.008}} &
{\pv{$<$.001}} \\
\addlinespace[2pt]

Instance B
& 0 & 76 & 72
& +40.1 & +7.3 & +42.1 & +29.8 \\[-2pt]
& & & &
{\pv{.008}} &
{\pv{.048}} &
{\pv{.008}} &
{\pv{$<$.001}} \\
\addlinespace[2pt]

Instance C
& 0 & 59 & 57
& +38.3 & +2.1 & +42.9 & +27.8 \\[-2pt]
& & & &
{\pv{.008}} &
{\pv{.587}} &
{\pv{.008}} &
{\pv{$<$.001}} \\
\addlinespace[2pt]

Instance D
& 0 & 75 & 71
& +24.2 & -0.1 & +31.9 & +18.7 \\[-2pt]
& & & &
{\pv{.008}} &
{\pv{1.000}} &
{\pv{.008}} &
{\pv{$<$.001}} \\

\midrule
\rowcolor{rowhl}
\textbf{Average}
& {--} & {--} & {--}
& \bfseries +34.5
& \bfseries +1.3
& \bfseries +39.8
& \bfseries +25.2 \\[-2pt]

\rowcolor{rowhl}
& & & &
{\pv{$<$.001}} &
{\pv{.551}} &
{\pv{$<$.001}} &
{\pv{$<$.001}} \\
\bottomrule
\end{tabular}
\end{table}

\paragraph{Evaluator sensitivity.}
Opus 4.6 serves as the primary LLM-as-a-judge throughout this study. To assess whether the observed critique-depth pattern depends on the choice of judge model, we repeat the comparison using Opus 4.6, Sonnet 4.6, and GPT-5.5. For each judge, we recompute extra divergence, defined as cross-depth divergence relative to the corresponding same-depth rerun variability.

\begin{figure}[htbp]
\centering
\includegraphics[width=0.7\linewidth]{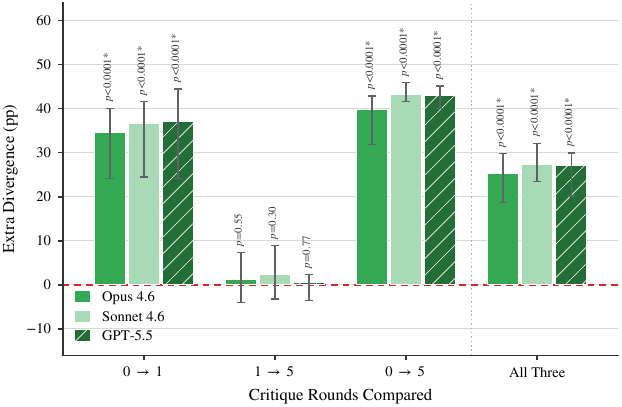}
\caption{\textbf{Sensitivity of extra divergence to the judge model.} Extra divergence is recomputed using Opus 4.6, Sonnet 4.6, and GPT-5.5 under the same evaluation criteria. Bars show extra divergence averaged across Instances, whiskers indicate the minimum-to-maximum range of the four instance-specific estimates, and annotations report two-sided permutation $p$ values. Across the three judges, extra divergence ranges from 34.5 to 37.3~pp for $K=0$ versus $K=1$ and from 39.8 to 43.3~pp for $K=0$ versus $K=5$, compared with 0.5 to 2.3~pp for $K=1$ versus $K=5$.}

\label{fig:sharedfamily}
\end{figure}

All three judges produce the same qualitative pattern (Figure~\ref{fig:sharedfamily}). The 0-to-1 comparison ranges from +34.5 to +37.3 pp of extra divergence, and the 0-to-5 comparison ranges from +39.8 to +43.3 pp; both comparisons are significant under every judge ($p<0.0001$). In contrast, the 1-to-5 comparison ranges from +0.5 to +2.3 pp. Thus, across these judges, most cross-depth separation appears when critique is first introduced, with limited additional separation from later rounds. 

\section{Pairwise Mechanism Comparison Methods}
\label{app:methods}

We compare three methods for determining whether a pair of hypotheses describes the same mechanism. Each method receives the same hypothesis statement, mechanistic explanation, and causal chain. In our TF--IDF implementation, vectors are constructed using sublinear term-frequency scaling and English stopword removal; pairs with cosine similarity of at least 0.25 are classified as the same mechanism. Dense embeddings are generated using \texttt{amazon.titan-embed-text-v2}, with pairs classified as the same mechanism when their cosine similarity is at least 0.77. This threshold is selected by matching the marginal same-pair rate to the range produced by the other methods, without using individual pair decisions as labels. The LLM-as-a-judge evaluator uses \texttt{claude-opus-4-6-v1} at temperature~0 and directly returns a binary same/different decision. The LLM-as-a-judge is queried three times per pair, including one presentation-order reversal, and the majority decision is used. The full presentation-order sensitivity is examined in Appendix~\ref{app:stability}, and evaluation rubric and its two prompt variants are in Appendix~\ref{app:prompts}.

\section{Controlled Perturbations for Grouping Decisions}
\label{app:variants}

\paragraph{Perturbation definitions.}
Each perturbation pair consists of one generated hypothesis and one edited variant designed to test how the comparison methods respond to a specific change. \textit{Wording Only} rewrites the mechanistic explanation while leaving the causal chain unchanged, whereas \textit{Wording + Chain} also rewrites the chain using audited biological aliases while preserving the ordered entities and causal relations. Both are intended to preserve the same mechanism and test invariance to meaning-preserving changes. \textit{First Step Swapped} replaces the initiating event while holding the downstream chain fixed and is treated as a different mechanism under the study rubric. \textit{Middle Step Swapped} replaces one intermediate step while preserving the initiating event and terminal finding.

\paragraph{Generation and structural validation.}
Variants are generated using \texttt{claude-sonnet-4-6} at temperature~0.3 and retained only if they pass deterministic validation against the original causal chain. \textit{Wording Only} must preserve all chain steps and regulatory directions, \textit{Wording + Chain} must preserve the ordered entities after alias normalization, and the step-swapped variants must change exactly one step at the designated position. Constructions that fail these checks are excluded. 

\section{Sensitivity Analyses for Pairwise Comparison Methods}
\label{app:stability}

\paragraph{Threshold sensitivity and cross-method agreement.}
At the selected thresholds of 0.25 for TF--IDF and 0.77 for embeddings, the three comparison methods achieve 85\%--89\% raw pairwise agreement and chance-corrected agreement of $\kappa\in[0.59,0.74]$. Sweeping the TF--IDF and embedding thresholds changes their same/different decisions, but no evaluated threshold produces complete agreement among the methods; the highest chance-corrected agreement with another method is $\kappa=0.74$. Thus, the observed differences among the methods are not attributable to a single threshold setting.

\paragraph{Judge-model and presentation-order sensitivity.}
On the 180 pairs drawn from the generated hypothesis set, repeating the full-rubric LLM judgment at temperature~0 in the same presentation order changes none of the decisions, whereas reversing the order of the two hypotheses changes 5 decisions (3\%). Repeating the comparison with three judge models yields 96.1\%--98.3\% pairwise agreement and $\kappa=0.86$--0.94. These results indicate high consistency across repeated evaluations and judge models, with a small residual dependence on presentation order.

\paragraph{Rubric and order interactions.}
We compare the full, step-by-step, and no-rubric prompts on the same  controlled perturbation pairs while holding the judge model, temperature, sampling procedure, and rendered inputs fixed (Figure~\ref{fig:judgeprompts}). Exact repetition in the same order changes no decisions under any prompt. Presentation-order reversal changes 7 decisions under each prompt, but the affected pairs differ: two are shared between the full and step-by-step rubrics, and none are shared with the no-rubric prompt.

\section{LLM-as-a-judge Prompts}
\label{app:prompts}

\subsection{Critique-engagement judge}

The critique-engagement judge decomposes each critique into individual demands and classifies the proposer's response to each demand as conceded, answered, partially addressed, or dropped.

\begin{promptbox}{Critique-engagement Judge}
\promptsize
\begin{Verbatim}
You are auditing one exchange from a scientific dispute. A critic agent has challenged a mechanistic hypothesis. The hypothesis' generator agent has answered. 

Judge ONLY whether the reply engages the specific challenge the critic agent made. Do NOT judge whether either side is scientifically right, whether the mechanism is plausible, or how well written either is.

For each attack's demand, say what the reply did with it, and label that demand with exactly one of:

  "answered"  -- the reply takes the demand on its own terms and gives a reason,
                 evidence, or a stated ground for refusing it.
  "conceded"  -- the reply gives the demand up: withdraws the claim, corrects it,
                 or downgrades it.
  "partial"   -- the reply engages this demand but leaves part of what was asked
                 for unmet (e.g. concedes the error but does not supply the
                 replacement, or argues the point without the evidence demanded).
  "dropped"   -- this demand is not taken up anywhere in the reply.

Return ONLY JSON, no prose:
{"demand_labels": ["answered"|"conceded"|"partial"|"dropped"]}

\end{Verbatim}
\end{promptbox}

\subsection{LLM-as-a-judge and rubric variants}

The LLM-as-a-judge determines whether two hypotheses describe the same or different mechanism. We present the full rubric used throughout the pairwise analyses, a step-by-step variant that records component-level judgments before the overall verdict, and a no-rubric variant that provides only the comparison task.

\begin{promptbox}{Full Rubric (used throughout)}
\promptsize
\begin{Verbatim}
You are auditing a set of mechanistic hypotheses. You will be shown TWO hypotheses, A and B, about the same instance and the same observed finding. Decide whether they propose the SAME underlying mechanism or DIFFERENT mechanisms. 

Call them the SAME mechanism when:
  - They name the same molecular initiating event (e.g. the same target, transporter, enzyme or pathway)
  - The essential route from that event to the observed finding is the same, allowing for one being described in more or less detail

Call them DIFFERENT mechanisms when any of these holds:
  - A different initiating event (e.g. receptor blockade vs. reactive metabolite formation vs. an immune response), even if downstream language overlaps
  - The same initiating event but a genuinely different load-bearing route to the observed finding
  - They assert opposite directions for the same step (one says a mediator rises where the other says it falls), or opposite orientation (A drives B vs. B drives A).

Differences in wording or naming conventions, the inclusion of additional downstream biomarkers itself, and the level of detail used to describe the same causal route do not constitute different mechanisms.

Return ONLY JSON, no prose:
{"verdict": "same"|"different"}
\end{Verbatim}
\end{promptbox}

\begin{promptbox}{Step-by-Step Rubric}
\promptsize
\begin{Verbatim}
You are auditing a set of mechanistic hypotheses. You will be shown TWO hypotheses, A and B, about the same instance and the same observed finding. Decide whether they propose the SAME underlying mechanism or DIFFERENT mechanisms. 

Call them the SAME mechanism when:
  - They name the same molecular initiating event (e.g. the same target, transporter, enzyme or pathway)
  - The essential route from that event to the observed finding is the same, allowing for one being described in more or less detail

Call them DIFFERENT mechanisms when any of these holds:
  - A different initiating event (e.g. receptor blockade vs. reactive metabolite formation vs. an immune response), even if downstream language overlaps
  - The same initiating event but a genuinely different load-bearing route to the observed finding
  - They assert opposite directions for the same step (one says a mediator rises where the other says it falls), or opposite orientation (A drives B vs. B drives A).

Differences in wording or naming conventions, the inclusion of additional downstream biomarkers itself, and the level of detail used to describe the same causal route do not constitute different mechanisms.

Answer three questions about the pair before giving your verdict, applying the
rules above to each one separately.
  1. INITIATING EVENT. Do A and B name the same molecular initiating event -- the
     same target, transporter, enzyme or pathway?
  2. ROUTE. Is the essential route from that event to the observed finding the
     same, allowing for one being described in more or less detail?
  3. TERMINAL OUTCOME. Do A and B end at the same observed finding?
Then give the overall verdict. 

Return ONLY JSON, no prose:
{"initiating_event": "same"|"different",
 "route": "same"|"different",
 "terminal_outcome": "same"|"different",
 "verdict": "same"|"different"}
\end{Verbatim}
\end{promptbox}

\begin{promptbox}{No-rubric}
\promptsize
\begin{Verbatim}
You are auditing a set of mechanistic hypotheses. You will be shown TWO hypotheses, A and B, about the same instance and the same observed finding. Decide whether they propose the SAME underlying mechanism or DIFFERENT mechanisms. 

Return ONLY JSON, no prose:
{"verdict": "same"|"different"}
\end{Verbatim}
\end{promptbox}

\end{document}